\documentclass[10pt,twocolumn,letterpaper]{article}

\usepackage{cvpr}              

\usepackage{booktabs}
\usepackage{multirow}
\usepackage{makecell}
\usepackage{verbatim}
\usepackage{graphicx}
\usepackage{float}
\usepackage{subcaption}
\usepackage{siunitx}
\usepackage{dblfloatfix}
\usepackage{array}
\definecolor{cvprblue}{rgb}{0.21,0.49,0.74}
\usepackage[pagebackref,breaklinks,colorlinks,allcolors=cvprblue]{hyperref}

\usepackage[normalem]{ulem}
\definecolor{gedit}{HTML}{FF0000}

\usepackage[accsupp]{axessibility}  
\def\paperID{*****} 
\def\confName{CVPR}
\def\confYear{2026}

\title{GABI: Geometry-Aware Boundary Integration for Spacecraft Segmentation}

\author{
\begin{tabular}{c}
Iason Georgios Velentzas \qquad Dhruv Ahuja \qquad Panagiotis Tsiotras \\
Georgia Institute of Technology \\
{\tt\small \{ivelentzas3,dahuja8,tsiotras\}@gatech.edu}
\end{tabular}
}

\begin{document}
\maketitle
\begin{abstract}
Accurate segmentation is 
crucial for autonomous spacecraft, as it directly affects downstream tasks related to 3D situational awareness. 
The harsh illumination conditions of space, however, produce images with high variability in appearance, 
hindering the generalization of
segmentation approaches 
across different spacecraft and environments. In this work, we propose GABI, a lightweight boundary-aware multi-task segmentation architecture that augments a convolutional backbone with an auxiliary distance-field prediction head. 
The distance field provides dense geometric supervision around object boundaries, encouraging the network to learn spatially consistent representations of spacecraft structures while maintaining low model complexity suitable for onboard perception systems.
We evaluated GABI against both an established convolutional baseline and a heavier transformer-based architecture. On the SPARK benchmark, distance-field supervision improves the baseline by up to $5\%$ in Average Precision while achieving performance comparable to the transformer models. In generalization experiments, GABI improves Average Precision by more than $50\%$ over the baseline. In cross-domain evaluation, the lightweight GABI variant performs within $5\%$ in IoU and F1-score of the heavier transformer model while being approximately ten times smaller. At the same time, the heavier GABI variant surpasses the transformer architectures while remaining nearly three times lighter.
\end{abstract}    
\section{Introduction}
\label{sec:intro}
Space exploration has long been at the forefront of technological advancement, with profound impacts on our everyday lives. From communication infrastructure and GPS, to weather forecasting, maintaining the sustainability of the space environment is critical to our society. Recent studies by the European Space Agency (ESA) ~\cite{esa_env_report, esa_health_index} have highlighted the urgent need for Active Debris Removal (ADR) and On-Orbit Servicing (OOS) due to increasing orbital congestion. Solving this problem requires that navigation in orbit be autonomous, efficient, robust, and at-scale~\cite{autonomy1}.

Monocular cameras are an attractive modality for scalability~\cite{visual1,visual2}. They provide rich information while meeting the weight, size, power, and cost requirements relative to other active sensing modalities. Moreover, advances in Deep Learning for perception have demonstrated remarkable robustness and efficiency. However, the scarcity of both synthetic and real-world images has limited the applicability of methods developed for space applications~\cite{scarcity}. The SPARK 2026 Challenge provides a dataset of 10 spacecraft with a variety of pose configurations, accompanied by masks for part segmentation. 
\begin{figure}[t]
\centering
\includegraphics[width=0.95\columnwidth]{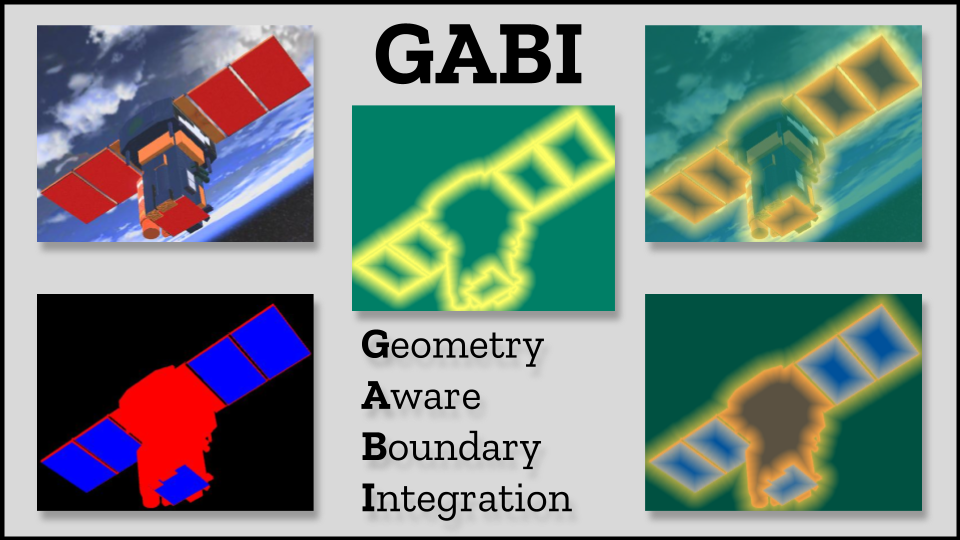}
\caption{GABI is using the distance field transform to integrate geometric information in the segmentation architecture}
\label{fig:teaser}
\end{figure}

Spacecraft segmentation is a critical task for spacecraft autonomy. Accurate localization and identification of space objects are crucial for many downstream tasks, such as pose estimation and 3D situational awareness~\cite{instance_seg1}. Target misidentification can lead to losing track of the target, wasting valuable fuel, and potentially failing the mission completely. Existing approaches to spacecraft segmentation address the challenge of developing efficient solutions, but robustness and generalization remain largely unexplored.

The space domain poses significant challenges for image processing algorithms~\cite{aouada}. Varying illumination conditions and the absence of atmospheric diffusion can produce images with markedly different photometric properties or lead to pronounced effects, such as sharp shadowing or lens flares. Similarly, the variety of image backgrounds, ranging from Earth at daytime or nighttime to the vast emptiness of space, is an important factor affecting the overall appearance of the image. Specifically for the predominantly texture-based task of spacecraft segmentation, the developed algorithms must be flexible and adapt to significantly different appearances of the same texture.

Despite the harsh illumination conditions in orbit, the geometric information in spacecraft imagery can help achieve robustness and in-domain generalization of the solution~\cite{nerf_geo}. Spacecraft geometry is typically characterized by rigid modular structures comprising planar or cylindrical surfaces with sharp edges. Despite the fact that structural elements in the images can appear overlapping and with fine detail, such as panel appendages or antennas, the observed spacecraft possess distinct geometric properties that can be leveraged as geometric priors or regularizers in perception to achieve robust target localization.

We propose \textbf{G}eometry-\textbf{A}ware \textbf{B}oundary \textbf{I}ntegration (\textbf{GABI}), a solution to the spacecraft segmentation problem that involves a geometric regularizer to achieve robust performance across datasets. We construct distance fields from existing segmentation masks and integrate them 
into a baseline efficient segmentation network. We formulate our solution as a multi-task learning approach, in which the geometric distance field serves as a regularizer for the learned features. The goal is for the network to learn features that capture both texture and geometric details, such as sharp straight edges or corners. Although the idea of explicitly modeling object boundaries for segmentation has been explored in different forms in the literature, learning a distance field provides a richer geometric representation. Instead of encoding only the location of the boundary, the distance field describes the spatial relationship of every pixel with the object contour, enabling the network to better capture the global structure of the spacecraft components.

This paper investigates the effect of learning geometry and texture together in the spacecraft segmentation context with the goal of robustness and generalization. We compare our proposed architecture with the baseline approach that we selected across different spacecraft and datasets. Given the lack of masks for different parts in other datasets, we conduct experiments of instance segmentation, which treats the whole spacecraft as foreground. Our experiments demonstrate that our network, while only marginally improving performance in the SPARK 2026 validation set, it is capable of producing significantly more complete foreground proposals in other datasets and across vastly different illumination conditions.

\section{Related Work}
\label{sec:formatting}

\paragraph{Efficient Semantic Segmentation}
Semantic segmentation aims to assign a semantic label to every pixel in an image, enabling dense scene understanding in autonomous navigation and remote sensing. Although early architectures demonstrated strong performance on dense pixel-wise prediction tasks, these same models rely on heavy backbone networks and intensive encoder-decoder structures~\cite{Long_2015_CVPR, badrinarayanan2017segnet, ronneberger2015u}. Mask R-CNN \cite{he2017mask}, for example, employs a two-stage approach that achieves strong instance segmentation performance but incurs a high computational cost, with approximately 44M parameters and a correspondingly high count of FLOPS in the ResNet-50-FPN backbone version~\cite{he2016deep}.

To enable efficient inference while maintaining strong representational capacity, lightweight backbone networks such as MobileNet~\cite{howard2017mobilenets} and EfficientNet~\cite{tan2019efficientnet} were introduced. Specifically, MobileNet achieves efficiency through depthwise separable convolutions, which factorize a standard convolution into a depthwise convolution followed by a pointwise $1\times1$ convolution. This decomposition dramatically reduces the computation and parameter count while preserving spatial feature extraction. EfficientNet extends this line of work by introducing a compound scaling strategy that scales network depth, width, and image resolution. 

These advances enabled the development of efficient segmentation architectures that leverage lightweight backbones with powerful segmentation heads~\cite{chen2017deeplab, xie2021segformer, paszke2016enet, poudel2019fast, mehta2018espnet}. Among these, DeepLabV3+~\cite{chen2017deeplab} has established itself as a widely adopted framework for its effective multi-scale context modeling via atrous spatial pyramid pooling (ASPP). The authors demonstrate that atrous convolution enables accurate multi-scale segmentation by expanding the receptive field through kernel dilation, without introducing additional learnable parameters. More recently, Segformer~\cite{xie2021segformer} introduced a hierarchical Mix Transformer (MiT) encoder with overlapping patch embeddings and spatial-reduction self-attention. The capacity to produce multi-scale feature maps, paired with a lightweight MLP decoder for feature fusion, enables efficient and accurate semantic segmentation.

\paragraph{Structure-Aware Image Segmentation}
A significant portion of the image segmentation literature has been devoted to coupling geometric and semantic information. The two main ways to use segmentation changes to infer geometry are either binary boundaries or continuous distance transforms.  

The works of~\citep{gated_scnn,semeda} utilize binary boundary maps to improve local segmentation close to ambiguous boundary predictions. Gated-SCNN~\cite{gated_scnn} proposed using the main and more powerful segmentation stream to guide a separate parallel stream that encodes shape information. The network's performance is also boosted via a novel duality regularization loss between the two streams. On the other hand, SEMEDA~\cite{semeda}  designed a semantic edge-aware loss in the feature space. These works achieve sharper predictions near the boundaries, boosting performance on thinner structures, without completely eliminating boundary misdetections.

%

Boundary-aware segmentation typically focuses on improving local accuracy near the boundary and on capturing high-frequency details where masks change. Distance fields, on the other hand, are used to add global context from the image and to exploit the structure of the scene even beyond the boundaries. Distance fields are a rich representation that encodes dense per-pixel boundary information, offering robustness to occlusions and noise.
The works of~\citep{aerial_DCNN, distance_regression} opt to learn a distance field for each label in the image. Learning signed distance fields~\cite{aerial_DCNN} for each label and possibly the corresponding class in a multi-task approach~\cite{aerial_DCNN} showcases the ability to capture spatial context in aerial images and a significant improvement in smoothness of predictions and lack of holes in the segmentation masks.

Similar to boundary approaches, many distance transform methods focus on enhancing predictions along the boundary.
%
The work of~\citep{boundary_instance} was motivated by the observation that the detect first and then segment approach of Mask R-CNN~\cite{maskrcnn} leads to unrecoverable segmentation errors when the bounding boxes are detected erroneously. The authors aimed to enable masks to extend beyond the initial proposed bounding boxes and introduced a distance-based mask representation that predicts a binary mask per class.
Investigating the distribution of mask prediction errors across baseline networks, the authors of~\cite{segfix} show that the mask predictions of the interior pixels are significantly more reliable. They propose a model-agnostic approach to segmentation refinement in which boundary pixels are localized by learning a boundary and direction map, and subsequently unstable predictions are replaced with the corresponding interior pixel.

%
In conclusion, the vast majority of works on structure-aware segmentation in terrestrial applications use distance fields and boundary masks to geometrically regularize the network prediction and enhance performance on fine-grained object boundaries. A commonly observed outcome is that pixel-wise losses often misconstrue the image topology, and adding extra geometric supervision helps to achieve structurally consistent mask predictions.

\paragraph{Spacecraft Segmentation}
In contrast to terrestrial applications, the space domain presents a substantially different set of challenges and characteristics. For spacecraft recognition, a typical categorization of interest is between the spacecraft body and panel material components, a convention also used in the SPARK 2026 challenge. Other datasets might involve more components, such as antennas~\cite{dt_ade}, or even focus on instance segmentation, where the need is to distinguish the spacecraft from the background~\cite{spe3r}. 

As the community addresses the problem of data scarcity and more datasets are generated, baseline segmentation algorithms are being evaluated as performance benchmarks
The authors of \citep{dt_ade} compared various state-of-the-art segmentation networks in both instance segmentation and parts recognition. The results of the latter experiment indicate that the baseline algorithms have trouble identifying the antennas, which are a complex and complex-shaped module, as well as facing problems when the antenna overlaps with the main body.
Prior work in~\citep{dt_china} proposes a semi-physical image collection system that focuses on the extreme effects of illumination conditions in space and on realistic, complex structures. The images have nine distinct annotated labels, and the results indicate that efficient segmentation architectures do not accurately capture small structures.
Recently,~\citep{dt_rice} have offered a new synthetic dataset and provided results for finetuning different versions of YOLO segmentation models. The authors formalized the onboard constraint requirements based on NASA standards and deemed that YOLOv8n and YOLOv11n are the best performing variants that meet these requirements~\cite{yolov8n, yolov11n}.


\paragraph{Structure-Aware Spacecraft Applications}
Despite the data scarcity in the space domain, there are some works that explore the intersection of geometric and semantic information of spacecraft. For the problem of line segment detection,~\citep{irolsd} used an illumination augmentation scheme, similar to typical homographic augmentation, to detect line segments robustly under various illumination conditions. The authors showed that the global context that is injected into the problem by using attraction fields~\citep{afm} is able to generate more complete line segments under harsh conditions, such as low brightness or low contrast. Moreover, IRoLSD~\cite{irolsd} exhibited remarkable generalization performance on unseen spacecraft geometries and environmental conditions in both synthetic and real-world images.

In particular for the problem of spacecraft segmentation,~\citep{wdicd} created a boundary and a distance field from the mask boundaries to augment the segmentation stream with two parallel heads: a boundary prediction head that learns the binary boundary mask and a head that uses the distance field contrastive context feature learning. This work focuses on improving segmentation locally across the boundary and, despite not learning a continuous distance field representation, it shows significant improvement in the overall boundary prediction. 

A more influential work in the field was presented in~\citep{spnv2}. Focused on instance segmentation, the authors explore the robustness of the segmentation pipeline using geometric primitives in a multi-task learning scenario. In addition to the segmentation head, the authors develop two heads, one for pose estimation and one for keypoint detection. The keypoints encode 2D geometric information, whereas the poses provide strong 3D projective-geometry information to the network. The complete evaluation suite employs various space-specific image augmentations, such as sun flare and style augmentations, and, respecting mission-operational constraints, it develops an online domain refinement module to assess performance gains in real-world scenarios.

Overall, prior work suggests that incorporating geometric representations alongside semantic predictions can improve the structural consistency of segmentation models. In this work, we investigate whether learning a continuous distance field can introduce a geometric inductive bias in the network, encouraging representations that remain robust across varying imaging conditions and unseen spacecraft geometries.

\section{Method}
\label{method}

\begin{figure*}[t]
    \centering
    \includegraphics[width=0.8\textwidth]{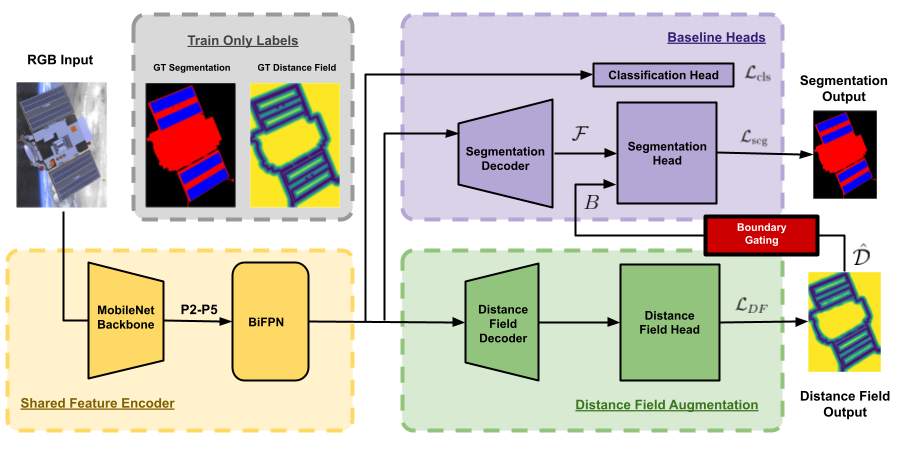}
    \caption{Overview of the proposed multi-task architecture. The predicted Distance Field is further used to generate a boundary-weighting map that modulates segmentation prediction, encouraging boundary-aware feature learning.}
    \label{fig:architecture}
\end{figure*}
This section presents a boundary-aware semantic segmentation framework developed for the SPARK 2026 Spacecraft Perception Challenge, which involves multi-task spacecraft perception tasks including classification, detection, and semantic segmentation. In our approach, semantic segmentation serves as the primary prediction task, while detection outputs are derived directly from the predicted segmentation masks via bounding box extraction. As a result, improvements in segmentation quality also directly impact detection performance.

To improve segmentation accuracy, particularly along object boundaries, we augment a standard encoder–decoder segmentation architecture with an auxiliary distance field prediction task. We first describe the baseline segmentation network used for the challenge. We then introduce the proposed distance field prediction module and describe its integration as an auxiliary head in the architecture. Finally, we present the training objectives used to jointly optimize segmentation and distance field supervision.

\subsection{Multi-Task Baseline Architecture}


The baseline architecture, depicted in Figure \ref{fig:architecture}, follows the encoder–decoder style architecture used in DeepLabV3+ and can be summarized by the components \textit{Shared Feature Encoder} and \textit{Baseline Heads}.
Feature extraction is performed using the lightweight MobileNetV2 backbone~\citep{howard2017mobilenets}, from which multiple resolution maps are extracted and fed as input to the \textit{Bidirectional Feature Pyramid Network (BiFPN)} ~\citep{tan2020efficientdet}. Projecting each level to a fixed channel dimension before fusion allows efficient cross-scale aggregation, combining detail-oriented features with semantic features.
The backbone network produces a pyramid of features $\text{P}2-\text{P}5$, with progressively lower spatial resolution. These features are fused using a \textit{BiFPN} module, which combines information across multiple scales through top-down and bottom-up pathways. The segmentation decoder follows the standard DeepLabV3+ architecture, aggregating multi-scale context via ASPP and fusing high-level semantic features with lower-level spatial features to generate pixel-wise predictions for the classes \textit{background}, \textit{spacecraft body}, and \textit{spacecraft panel}.

In our design, 
spacecraft classification is performed using a global pooling on the coarser output of \textit{BiFPN} feature map, followed by dropout and a linear layer to predict the spacecraft class. The shared head structure enables efficient joint-learning of semantic features related to spacecraft recognition. The segmentation head and classification head are supervised simultaneously, with both tasks monitored using cross-entropy loss.

\subsection{Boundary Integration}

While binary segmentation masks define a discontinuous indicator function in the image domain, the distance transform defines a continuous scalar field that encodes the distance to the nearest object boundary~\cite{afm}. The distance field is computed from mask boundary pixels where label transitions occur. For each pixel in the image domain, we compute the Euclidean distance to the nearest mask boundary and clip the value at a maximum distance $d_{\max}$.

\paragraph{Distance Field Implementation}
The network predicts a bounded distance field $\hat{D}$ by upscaling the encoder-generated features through a series of convolutional layers and bilinear interpolation, which are followed by ReLU activations and Batch normalization. 
Similarly to DeepLSD's line neighborhood, we only use a region around the boundary as valid for distance field learning.

\paragraph{Boundary Gating for Segmentation}
To utilize the geometric information encoded from the boundary and improve segmentation performance along the boundary, we propose to use a boundary-gating block that modulates segmentation features using the predicted distance field.
More specifically, let $\mathbf{F}$ denote the segmentation decoder feature map. Using the predicted distance field $\hat{D}$, we construct a gating map
$B = w \exp\!\left(-\frac{\hat{D}^2}{2\sigma^2}\right)$,
where $w$ and $\sigma$ are hyperparameters controlling the strength of the gating signal and the decay rate of the Gaussian weighting, respectively.
The gated feature representation is then computed as
\begin{equation}
\mathbf{F}_{\text{gate}} =
\mathbf{F} \odot (1 + \alpha \mathbf{B}),
\end{equation}
where $\odot$ denotes element-wise multiplication and $\alpha$ controls the strength of the boundary modulation. This mechanism is similar to the Gated-SCNN attention map, such that it amplifies feature responses near predicted object boundaries, encouraging the segmentation decoder to focus on high-frequency structural cues while preserving global contextual information~\cite{gated_scnn}.

\subsection{Loss Functions}

The multi-task perception objective function can be expressed as a combination of the distance field, segmentation, and classification losses, which are described below:

\begin{equation}
\mathcal{L}_{\text{total}} =
\lambda_{\text{DF}}\mathcal{L}_{\text{DF}}
+
\lambda_{\text{seg}}\mathcal{L}_{\text{seg}}
+
\lambda_{\text{cls}}\mathcal{L}_{\text{cls}}
\end{equation}

\paragraph{Classification Loss}
For $\mathcal{C}$ spacecraft models with $y_{\mathcal{C}}$ ground-truth labels, the
spacecraft classification loss is defined as the cross-entropy loss on $p_{\mathcal{C}}$:

\begin{equation}
\mathcal{L}_{\text{cls}} =
-\sum_{c=1}^{C} y_c \log p_c ,
\end{equation}

\paragraph{Distance Field Loss}

The distance field prediction is trained using a regression loss between the normalized distance field prediction $\hat{D}_n$ and the normalized ground-truth distance field $D_n$, around a valid neighborhood region of size $ r\in\mathbb{R}_{+}$.

\begin{equation}
   \mathcal{L}_{\text{DF}}= \lVert \hat{\mathcal{D}}_n - \mathcal{D}_n\rVert_1, \quad \mathcal{D}_n = -\log \left( \frac{\mathcal{D}}{r}\right)
\end{equation}

\paragraph{Segmentation Loss}

The segmentation objective is defined as a standard pixel-wise cross-entropy loss applied over the image domain $\Omega$ and the predicted probability of the ground-truth segmentation class $p_{S_{\text{gt}}}$:

\begin{equation}
\mathcal{L}_{\text{seg}} =
-\frac{1}{|\Omega|}
\sum_{\Omega}
\log
p_{S_{\text{gt}}},
\end{equation}

\section{Experiments}
\label{method}

In this section, we evaluate the performance of the proposed architecture against different levels of generalizability. First, we evaluated the segmentation performance for every mask within the SPARK dataset. Second, we withhold part of the dataset and test the ability of the network to generalize in unseen spacecraft geometries within the same dataset domain. Finally, we evaluate the generalization of GABI across other synthetic domains. With these experiments, we show that distance-field supervision improves structural consistency beyond standard pixel-wise segmentation performance.

\subsection{Experimental Setup}
\paragraph{Datasets}
We used the SPARK 2026 Challenge dataset for the primary segmentation task. It is generated with Unreal Engine with ten distinct spacecraft models of varying scales, viewpoints, backgrounds, and lighting conditions. Each model contains 6,000 images for training and 2,000 images for evaluation. A subset of the dataset includes challenging visual effects, such as intense lens flares that introduce additional variability in the appearance of the models.
For the domain-generalization task, we used the SPE3R dataset~\cite{spe3r}, which contains 100k synthetic images of 100 spacecraft models. SPE3R exhibits greater variation in brightness and contrast, whereas in SPARK images, the illumination profile is more homogeneous. Furthermore, images of SPE3R have prominent self-shadowing effects, rendering portions of the spacecraft nearly indistinguishable from the background.
The two datasets differ both in spacecraft geometries and in illumination properties, allowing a meaningful cross-dataset generalization experiment.

\paragraph{Evaluation metrics}
Segmentation performance is evaluated using standard region-based metrics, for instance, and semantic segmentation. We report Average Precision (AP) at multiple Intersection-over-Union (IoU) thresholds. The evaluation metric of the SPARK 2026 Challenge for the segmentation task is the mean Average Precision, calculated by averaging AP scores across IoU thresholds from 0.50 to 0.95 with a step size of 0.05. This metric provides a comprehensive evaluation across multiple levels of spatial overlap. In addition, we report the standard evaluation metrics $\text{AP}_{50}$ and $AP_{75}$, and when meaningful, the even stricter threshold of $\text{AP}_{90}$.
For experiments conducted on the SPARK dataset, we evaluated the segmentation performance separately for the spacecraft body and solar panel components. For cross-domain experiments, the evaluation is performed in the foreground extraction task, which combines body and panel predictions. Since the spacecraft occupies a large portion of the image, it is deemed appropriate to also report the F1-score, predicated on precision and recall metrics to avoid misinterpreting high IoU that stem from network's over-prediction.

\begin{table*}[t]
\centering
\small
\setlength{\tabcolsep}{4pt}
\renewcommand{\arraystretch}{0.85}
\caption{Segmentation performance comparison on SPARK dataset. For each class (body, panel), we report AP at IoU thresholds of 0.50, 0.75, and 0.90, along with mAP averaged over IoU thresholds from 0.50 to 0.95 with a step size of 0.05.}
\label{tab:exp1}
\begin{tabular}{lccccccccccc}
\toprule
\multirow{2}{*}{Model} & \multirow{2}{*}{Params} & \multirow{2}{*}{FLOPs} 
& \multicolumn{4}{c}{Body} 
& \multicolumn{4}{c}{Panel} 
 \\
\cmidrule(lr){4-7} \cmidrule(lr){8-11}
& & & $\text{AP}_{50}$ & $\text{AP}_{75}$ & $\text{AP}_{90}$ & mAP & $\text{AP}_{50}$ & $\text{AP}_{75}$ & $\text{AP}_{90}$ & mAP \\
\midrule
BL-v3s        & 2.7 M & 4.8 G & \textbf{99.8} & 94.1 & 74.2 & 88.5 & 80.2 & 73.6 & 53.1 & 68.6   \\
GABI-v3s      & 2.8 M & 8.6 G & \textbf{99.8} & 94.3 & 74.5 & 89.2 & 84.3 & 76.2 & 56.9 & 73.1   \\
BL-v2       & 3.9 M & 20.5 G & \textbf{99.8} & \underline{96.1} & \underline{81.8} & \underline{91.4}  &82.3  &77.0  &61.3  & 71.8   \\
GABI-v2     & 4.2 M & 34.5 G & \textbf{99.8} & \textbf{96.5}  & \textbf{82.3} & \textbf{91.9} & \underline{84.5} & \underline{79.8} & \underline{66.4}  & \underline{76.1}  \\
SegFormer-b0    & 3.7 M & 44.0 G & 99.6 & 89.9  &  62.4& 74.1 & 84.4 & 78.7 &65.9  &74.8   \\
SegFormer-b1    & 13.7 M & 86.6 G & 99.7 & 90.1 & 64.1 & 85.2 & \textbf{86.1} & \textbf{80.4}  & \textbf{68.9}  & \textbf{77.2} \\
\bottomrule
\end{tabular}
\end{table*}
\begin{table*}[t]
\centering
\small
\setlength{\tabcolsep}{4pt}
\renewcommand{\arraystretch}{0.85}
\caption{Segmentation performance comparison on hidden spacecraft of SPARK dataset. For each spacecraft (Soho, Proba3ocs) and for each class (body, panel), we report AP at IoU thresholds of 0.50 and 0.75, together with mAP averaged over IoU thresholds from 0.50 to 0.95 with a step size of 0.05.}
\label{tab:exp2}
\begin{tabular}{lcccccccccccc}
\toprule
\multirow{2}{*}{Model}
& \multicolumn{3}{c}{Soho -  Body}
& \multicolumn{3}{c}{Soho - Panel}
& \multicolumn{3}{c}{Proba3ocs - Body}
& \multicolumn{3}{c}{Proba3ocs - Panel} \\
\cmidrule(lr){2-4}
\cmidrule(lr){5-7}
\cmidrule(lr){8-10}
\cmidrule(lr){11-13}
& $\text{AP}_{50}$ & $\text{AP}_{75}$ & mAP
& $\text{AP}_{50}$ & $\text{AP}_{75}$ & mAP
& $\text{AP}_{50}$ & $\text{AP}_{75}$ & mAP
& $\text{AP}_{50}$ & $\text{AP}_{75}$ & mAP \\
\midrule
BL-v3s        & 71.5 & 20.8 & 29.3 & 05.9 & 00.2 &01.4 & 96.4 & 89.4 &82.0 & 74.1 & 70.1 & 60.8\\
GABI-v3s      & \underline{86.1} & \underline{35.9}  & \underline{42.3} &  25.9 & 10.4 & 12.3 & \underline{98.9} & 91.1 & 83.9 &\underline{81.3} & \underline{78.1} & \underline{69.1} \\
BL-v2       &  82.7 & 35.8   &36.4  & 24.7 & 12.0 &11.7  &97.3 & \underline{91.9} & \underline{84.2} &75.0  &72.4  & 64.4 \\
GABI-v2     & \textbf{89.2} & \textbf{39.5} & \textbf{45.8} &  \textbf{38.7} & \textbf{26.2} &\textbf{24.3 }&\textbf{99.2 }&\textbf{93.8} & \textbf{88.2}& \textbf{82.1}& \textbf{79.1 }& \textbf{74.4}\\
SegFormer-b0  &67.7  & 17.5 & 26.9 &  19.6 & 07.0 & 08.8 &98.4& 89.7 & 82.0 & 71.7  &68.0 & 58.7 \\
SegFormer-b1  & 68.8 & 25.0 & 31.2 &\underline{31.9}& \underline{18.4} & \underline{17.5} & 98.7 & 90.6 & 83.2 & 71.4 & 68.5 & 58.9\\
\bottomrule
\end{tabular}
\end{table*}

\begin{figure*}[!t]
    \centering
    \setlength{\tabcolsep}{0.7pt} 

    \begin{subfigure}[t]{0.495\textwidth}
        \centering
        \begin{tabular}{@{}ccccc@{}}
            \small \textbf{GT} &
            \small \textbf{BL} &
            \small \textbf{SegFormer} &
            \small \textbf{DF} &
            \small \textbf{GABI} \\

            \includegraphics[width=0.20\linewidth]{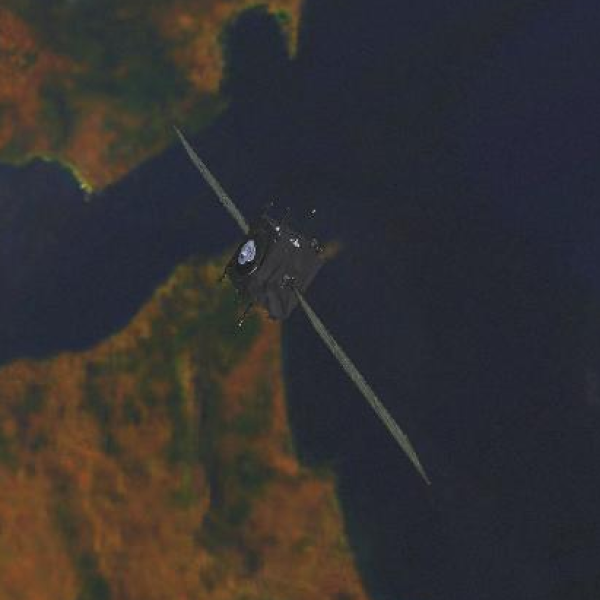} &
            \includegraphics[width=0.20\linewidth]{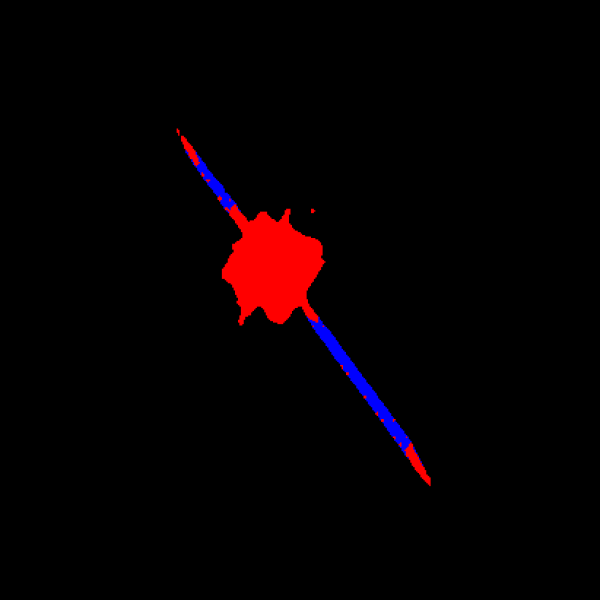} &
            \includegraphics[width=0.20\linewidth]{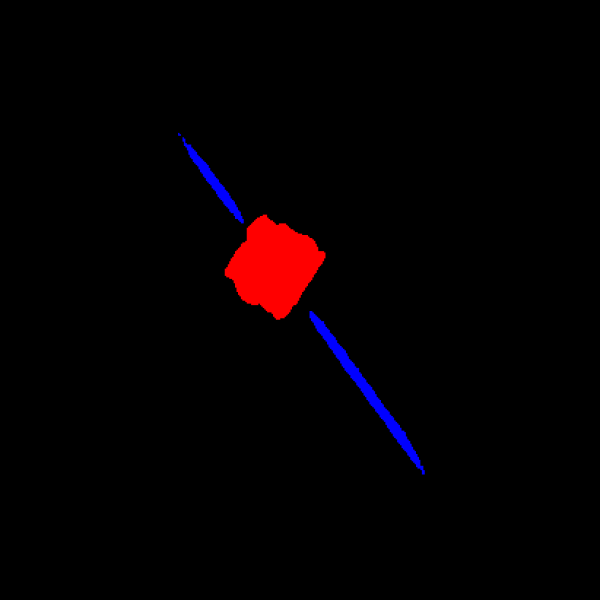} &
            \includegraphics[width=0.20\linewidth]{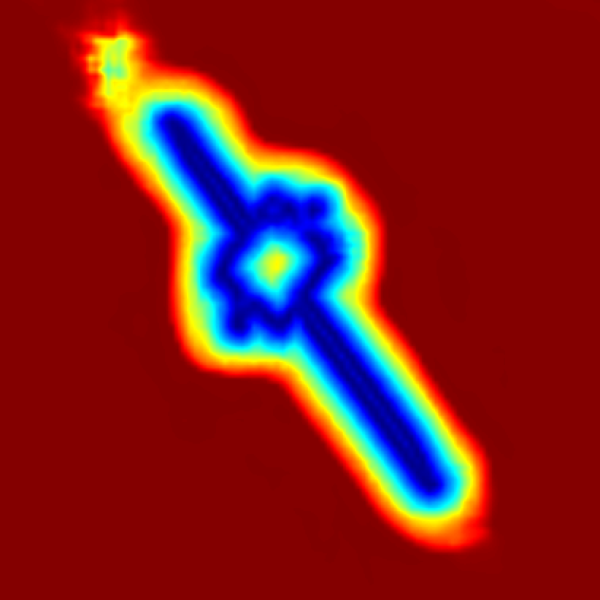} &
            \includegraphics[width=0.20\linewidth]{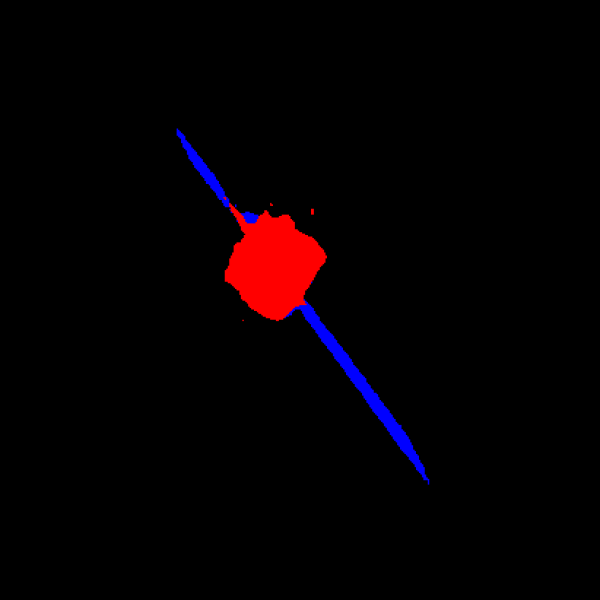} \\

            \includegraphics[width=0.20\linewidth]{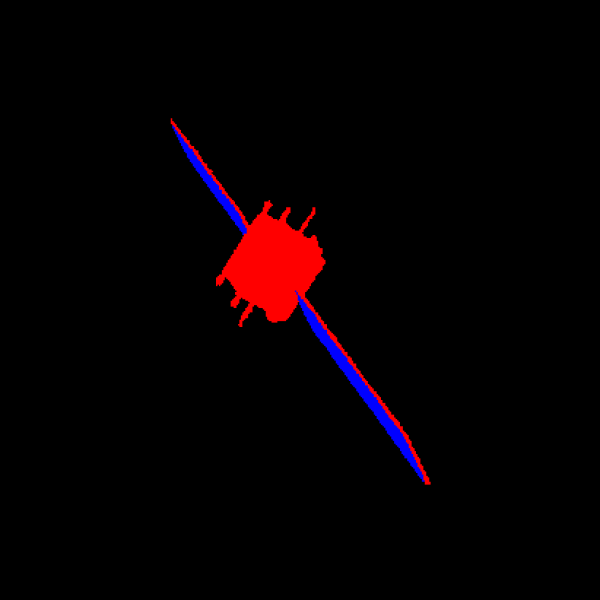} &
            \includegraphics[width=0.20\linewidth]{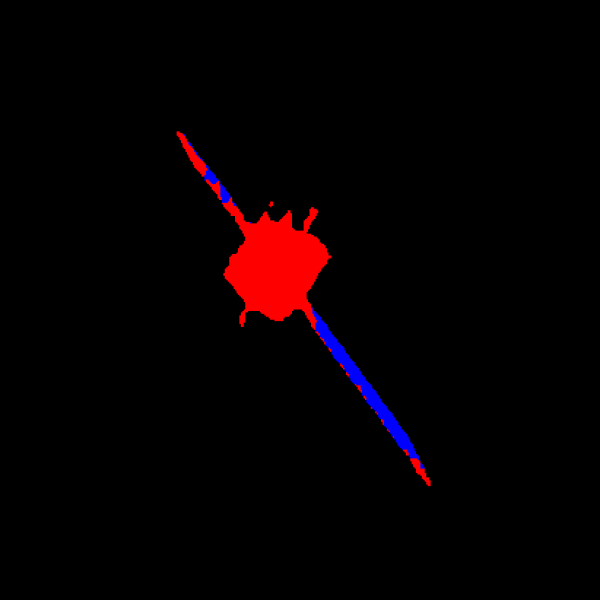} &
            \includegraphics[width=0.20\linewidth]{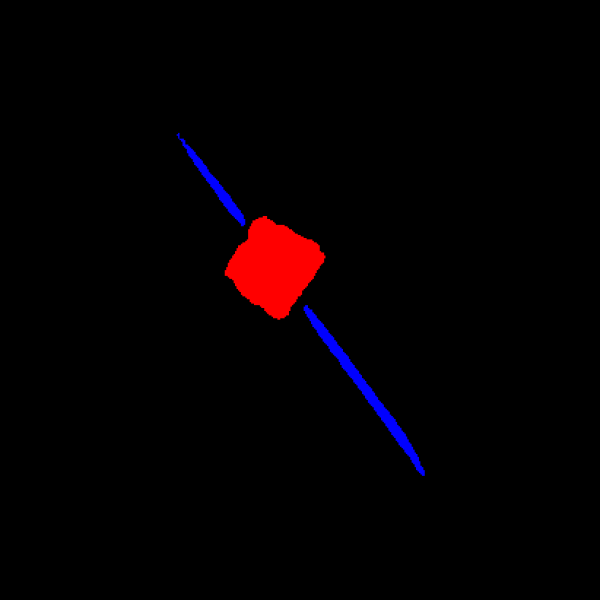} &
            \includegraphics[width=0.20\linewidth]{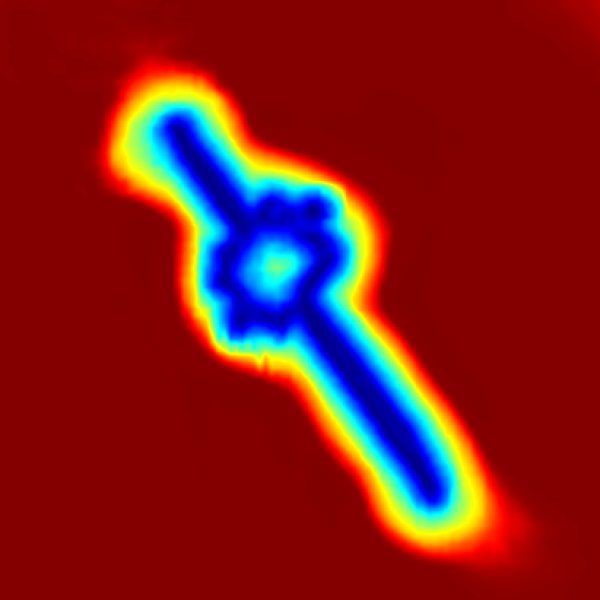} &
            \includegraphics[width=0.20\linewidth]{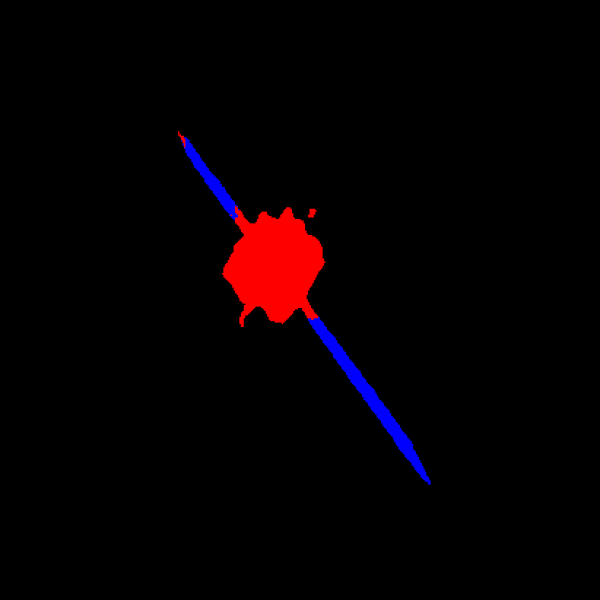} \\
        \end{tabular}
        \caption{Multi-task experiment.}
        \label{fig:exp1}
    \end{subfigure}
    \hfill
    \begin{subfigure}[t]{0.495\textwidth}
        \centering
        \begin{tabular}{@{}ccccc@{}}
            \small \textbf{Input} &
            \small \textbf{BL} &
            \small \textbf{SegFormer} &
            \small \textbf{DF} &
            \small \textbf{GABI} \\

            \includegraphics[width=0.20\linewidth]{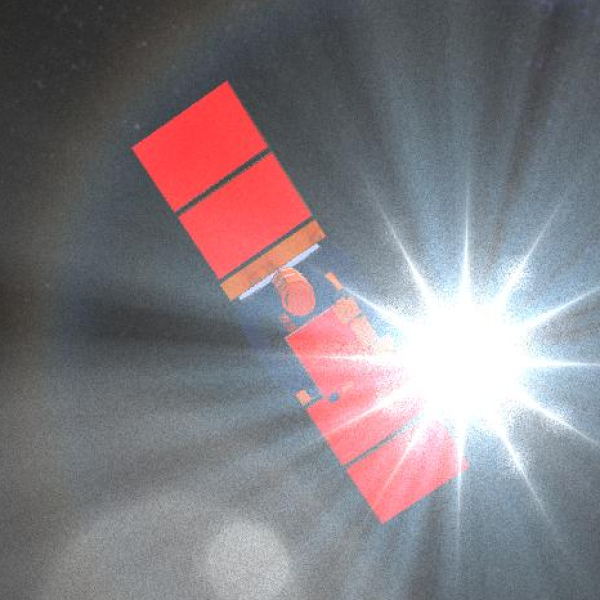} &
            \includegraphics[width=0.20\linewidth]{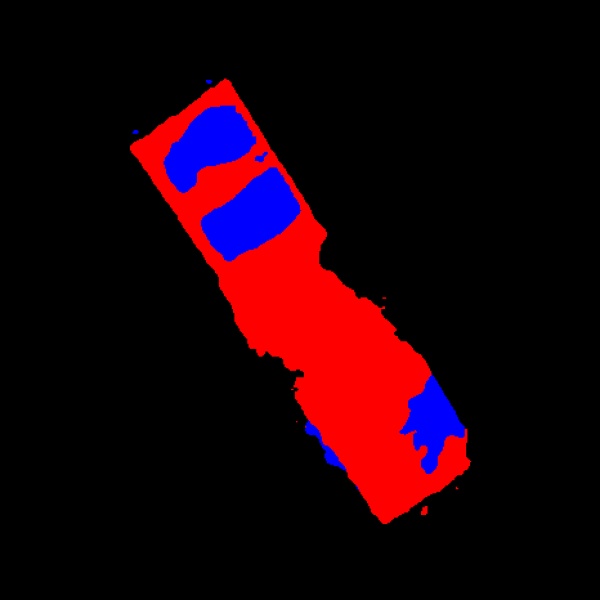} &
            \includegraphics[width=0.20\linewidth]{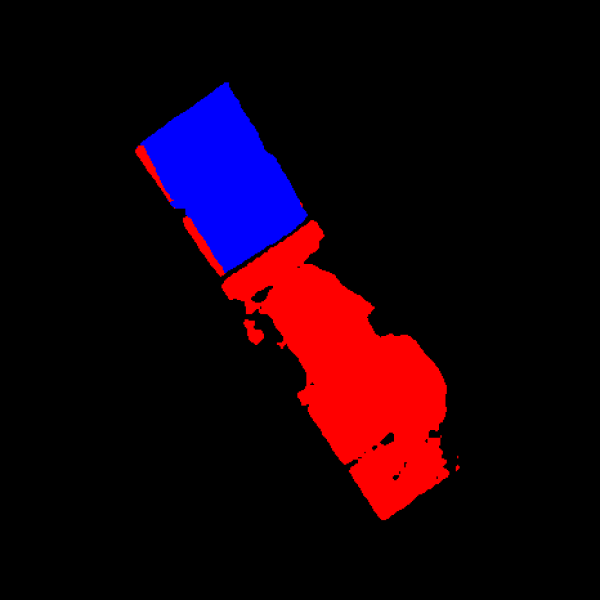} &
            \includegraphics[width=0.20\linewidth]{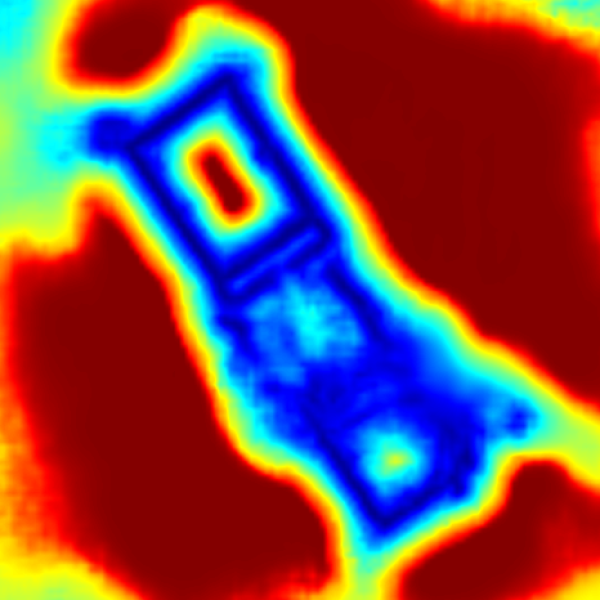} &
            \includegraphics[width=0.20\linewidth]{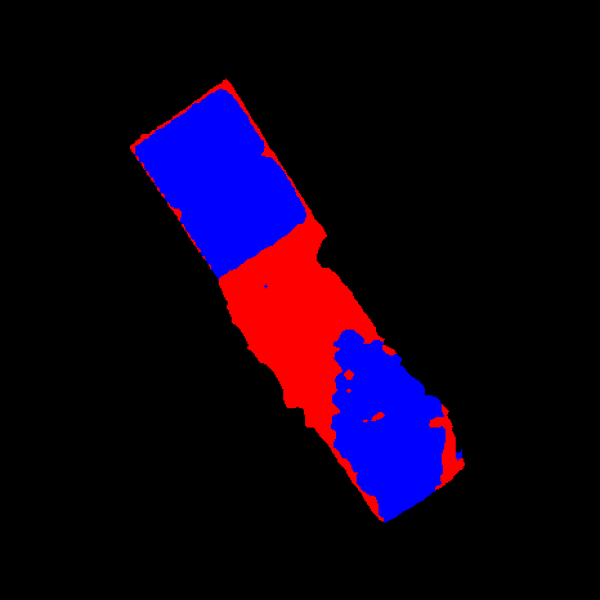} \\

            \includegraphics[width=0.20\linewidth]{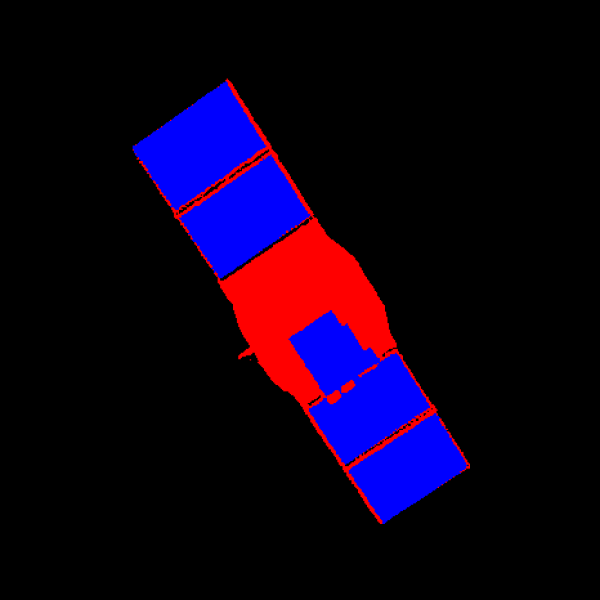} &
            \includegraphics[width=0.20\linewidth]{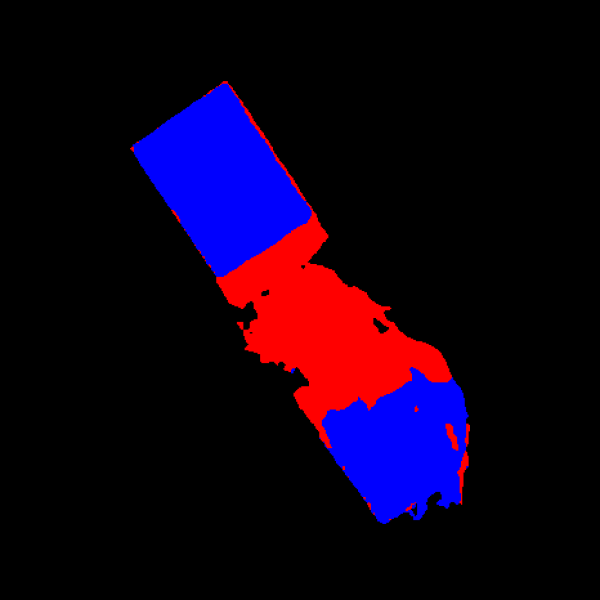} &
            \includegraphics[width=0.20\linewidth]{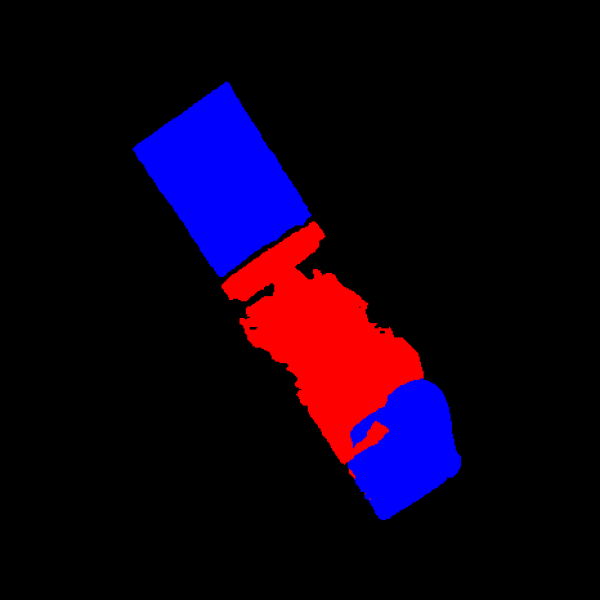} &
            \includegraphics[width=0.20\linewidth]{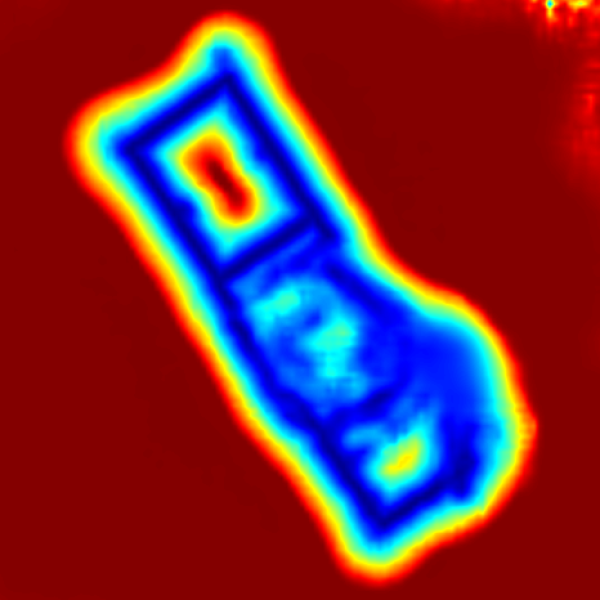} &
            \includegraphics[width=0.20\linewidth]{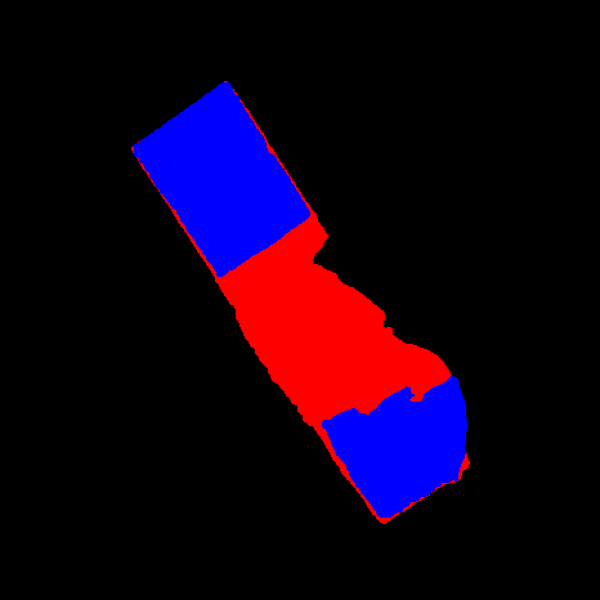} \\
        \end{tabular}
        \caption{Generalization in SPARK experiment on the Soho spacecraft.}
        \label{fig:exp2}
    \end{subfigure}

    \caption{Qualitative comparison of BL, SegFormer, and GABI. In both subfigures, the first column contains the input image and the ground-truth mask. Columns 2--3 show the segmentation predictions of BL and SegFormer~\cite{xie2021segformer}, while columns 4--5 show the distance field and segmentation prediction of GABI, respectively. The top row corresponds to the light models (BL-v3s, GABI-v3s and SegFormer-b0), while the bottom row corresponds to the heavier models (BL-v2, GABI-v2 and SegFormer-b1).}
    \label{fig:qualitative_comparison}
\end{figure*}

\begin{table*}[t]
\centering
\small
\caption{Segmentation performance comparison on three spacecraft models of SPE3R~\cite{spe3r} dataset. For each spacecraft (Apollo, Grace, lro), we report AP at IoU thresholds of 0.50 and 0.75, together with mAP averaged over IoU thresholds from 0.50 to 0.95 with a step size of 0.05. We provide the F1-scores and mean IoU over all images of the spacecraft.}
\label{tab:exp3}
\setlength{\tabcolsep}{4pt}
\renewcommand{\arraystretch}{0.85}

\begin{tabular}{lccccccccccccccc}
\toprule
\multirow{2}{*}{Model} &
\multicolumn{5}{c}{Apollo} &
\multicolumn{5}{c}{Grace} &
\multicolumn{5}{c}{lro} \\
\cmidrule(lr){2-6}
\cmidrule(lr){7-11}
\cmidrule(lr){12-16}
 & $\text{AP}_{50}$ & $\text{AP}_{75}$& mAP & $\text{F}_1$ & IoU
 & $\text{AP}_{50}$ & $\text{AP}_{75}$ & mAP& $\text{F}_1$ & IoU
 & $\text{AP}_{50}$ & $\text{AP}_{75}$ & mAP & $\text{F}_1$ & IoU \\
\midrule
BL-v3s &  60.4& 16.2 &  25.1& 67.7 & 54.0 & 69.5 & 52.3 & 49.1 & 72.1 &78.7  & 73.1  &35.4  & 38.5 & 74.0 & 62.1 \\
GABI-v3s &  \underline{92.1}& 65.2 & 54.4 &\underline{84.6}  &75.2  & \underline{90.0} & \underline{78.7}& 68.7 &\underline{88.2}  &\underline{80.9}  & 89.7 & 65.3 & 56.9 & 84.5 &75.0  \\
BL-v2 &  67.0& 19.5 & 29.2 & 70.9 & 57.6 & 75.8 & 62.2 & 58.9 & 78.7 & 71.1 &  75.4& 47.7  & 46.3  & 76.1 & 64.7 \\
GABI-v2 & \textbf{94.9} & \textbf{71.9} &\underline{58.6} & \textbf{86.8} & \textbf{76.8} &  \textbf{95.9}& \textbf{86.8} & \textbf{74.2} &\textbf{89.4}  &\textbf{83.9 } & \textbf{96.2} & \textbf{87.7} & \textbf{69.1} & \textbf{89.4} &\textbf{ 81.2 } \\
SegFormer-b0 & 86.5  & 54.2 & 51.9 & 82.5 & 72.2 & 84.2 & 72.9 & 68.3 & 84.6 & 78.2 & 90.6& 62.7 & 60.9 & 86.4 & 77.5 \\
SegFormer-b1 & 89.4 & \underline{67.4} & \textbf{60.9} & 85.5 & \textbf{76.8} & 84.5 &  73.2&  \underline{69.1}& 84.7 & 78.5 & \underline{92.7} & \underline{74.9} & \underline{66.1} & \underline{87.2} & \underline{79.4} \\
\bottomrule
\end{tabular}
\end{table*}
\begin{table*}[t]
\centering
\small
\caption{Segmentation performance comparison on three spacecraft models of SPE3R~\cite{spe3r} dataset. For each spacecraft (Apollo, Grace, lro), we report AP at IoU threshold of 0.75, together with mAP averaged over IoU thresholds from 0.50 to 0.95 with a step size of 0.05. We additionally provide the F1-scores and mean IoU over all images of the spacecraft.}
\label{tab:exp3}
\setlength{\tabcolsep}{4pt}
\renewcommand{\arraystretch}{0.85}

\begin{tabular}{lcccccccccccc}
\toprule
\multirow{2}{*}{Model} &
\multicolumn{4}{c}{Apollo} &
\multicolumn{4}{c}{Grace} &
\multicolumn{4}{c}{lro} \\
\cmidrule(lr){2-5}
\cmidrule(lr){6-9}
\cmidrule(lr){10-13}
& $\text{AP}_{75}$ & mAP & $\text{F}_1$ & IoU
& $\text{AP}_{75}$ & mAP & $\text{F}_1$ & IoU
& $\text{AP}_{75}$ & mAP & $\text{F}_1$ & IoU \\
\midrule

BL-v3s 
& 16.2 & 25.1 & 67.7 & 54.0 
& 52.3 & 49.1 & 72.1 & 78.7  
& 35.4 & 38.5 & 74.0 & 62.1 \\

GABI-v3s 
& 65.2 & 54.4 & \underline{84.6} & 75.2  
& \underline{78.7} & 68.7 & \underline{88.2} & \underline{80.9}  
& 65.3 & 56.9 & 84.5 & 75.0  \\

BL-v2 
& 19.5 & 29.2 & 70.9 & 57.6 
& 62.2 & 58.9 & 78.7 & 71.1 
& 47.7 & 46.3 & 76.1 & 64.7 \\

GABI-v2 
& \textbf{71.9} & \underline{58.6} & \textbf{86.8} & \textbf{76.8} 
& \textbf{86.8} & \textbf{74.2} & \textbf{89.4} & \textbf{83.9} 
& \textbf{87.7} & \textbf{69.1} & \textbf{89.4} & \textbf{81.2} \\

SegFormer-b0 
& 54.2 & 51.9 & 82.5 & 72.2 
& 72.9 & 68.3 & 84.6 & 78.2 
& 62.7 & 60.9 & 86.4 & 77.5 \\

SegFormer-b1 
& \underline{67.4} & \textbf{60.9} & 85.5 & \textbf{76.8} 
& 73.2 & \underline{69.1} & 84.7 & 78.5 
& \underline{74.9} & \underline{66.1} & \underline{87.2} & \underline{79.4} \\
\bottomrule
\end{tabular}
\end{table*}
\paragraph{Implementation details}
As mentioned above, we compare the performance of GABI against its baseline (BL), which uses the same architecture but without the distance-field augmentation. To ensure a fair comparison, the baseline model and the proposed architecture share the same backbone, training procedures, and optimization settings. Only the distance-field prediction head and its associated loss function are unique to GABI.
We compare two variants of each model, \textit{v2} and \textit{v3s}, which are named after the mobilenet backbone that they use. In addition to these models, we include the two lightest versions of SegFormer~\citep{xie2021segformer}, \textit{b0} and \textit{b1}. SegFormer is a transformer-based segmentation network with a relatively lightweight architecture that has demonstrated strong performance and robustness across multiple segmentation benchmarks. Including this model allows us to compare GABI not only against its direct convolutional baseline, but also against a competitive architecture with strong generalization capabilities.

To preserve detail, we train and test with full size images (1024x1024). We use the AdamW optimizer with the initial learning rate set to $3 \times 10^{-4}$ and the weight decay set to $1 \times 10^{-2}$. Training is performed by finetuning pretrained models for 20 epochs using a batch size of 10 images on RTX-5090 GPUs.

\subsection{Multi-Task Experiment} 
\label{sec:exp1}
~\cref{tab:exp1} presents the segmentation results for the two classes averaged on all spacecraft of the SPARK dataset. Overall, GABI consistently outperforms the baseline configurations with minimal parameter overhead. For the panel class, which consists of smaller, finely detailed structures, the performance increase is more than $5\%$ and is mainly observed at higher IoU thresholds, which denotes better segmentation accuracy. The fact that performance on the body class is satisfactory in both convolutional approaches can be explained by the homogeneity of the SPARK dataset in terms of appearance of texture, with limited variability in contrast and shadowing. The marginal improvement we observe in GABI corroborates the conclusion that the performance enhancement is due to the distance field's ability to encode clear geometry and better represent boundaries, as qualitatively observed also in~\cref{fig:exp1}. 

Regarding SegFormer architectures, we observe that performance on the body class is significantly lower at higher IoU thresholds than with CNN-based architectures. This behavior is consistent with previous studies showing that CNNs benefit from strong spatial inductive biases such as locality and translation equivariance, allowing them to learn geometric structures efficiently from limited data, whereas transformer architectures require larger datasets to learn these priors implicitly \cite{dosovitskiy2020image, d2021convit}. This is also visible in ~\cref{fig:exp1}, in which the SegFormer prediction fully captures the semantic meaning of panels in the image, but fails to add details in the spacecraft body prediction.

\begin{figure*}[t]
    \centering
    \setlength{\tabcolsep}{0.7pt}
    \begin{tabular}{cccccccc}
        \small \textbf{RGB} &
        \small \textbf{Mask} &
        \small \textbf{BLv2} &
        \small \textbf{SegFormer-b1} &
        \small \textbf{GABIv3s-DF} &
        \small \textbf{GABIv3s} &
        \small \textbf{GABIv2-DF} &
        \small \textbf{GABIv2} \\

        \includegraphics[width=0.11\textwidth]{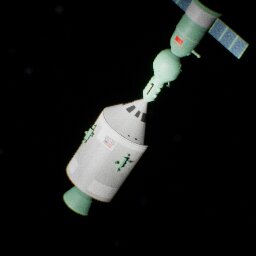} &
        \includegraphics[width=0.11\textwidth]{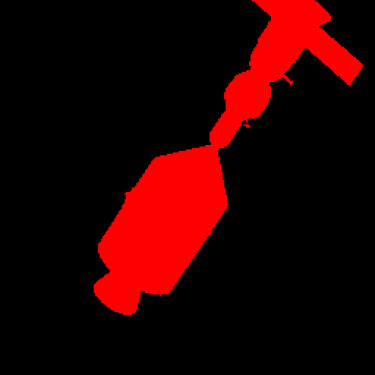} &
        \includegraphics[width=0.11\textwidth]{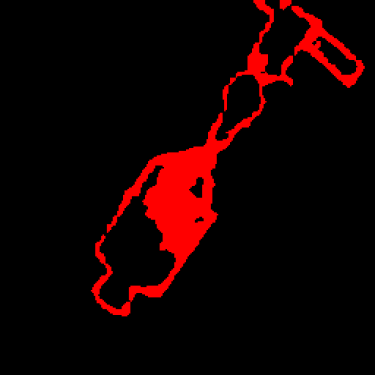} &
        \includegraphics[width=0.11\textwidth]{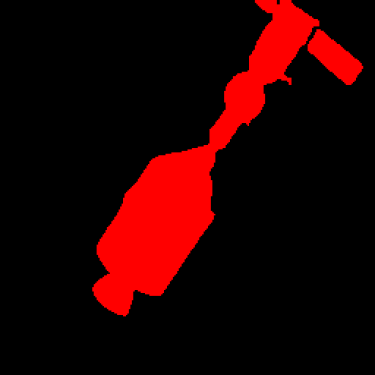} &
        \includegraphics[width=0.11\textwidth]{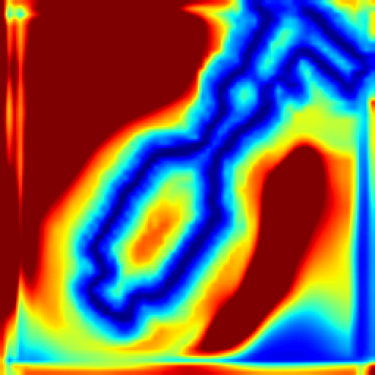} &
        \includegraphics[width=0.11\textwidth]{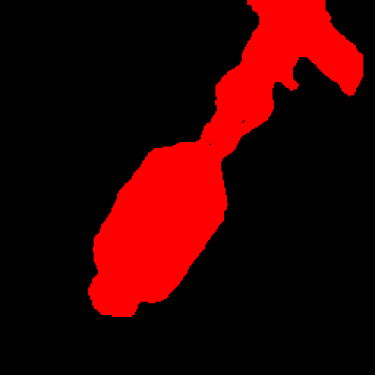} &
        \includegraphics[width=0.11\textwidth]{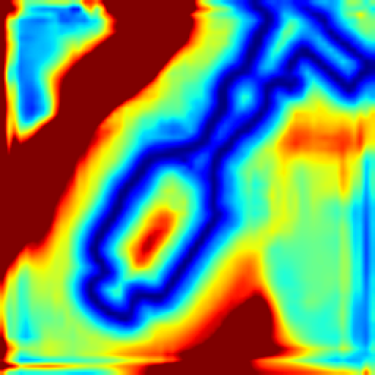} &
        \includegraphics[width=0.11\textwidth]{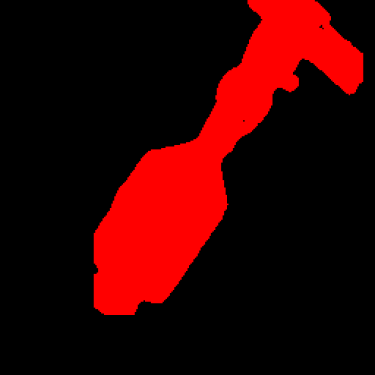} \\

        \includegraphics[width=0.11\textwidth]{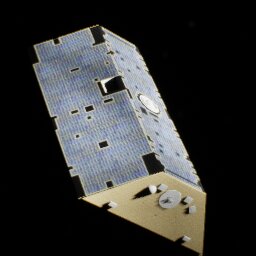} &
        \includegraphics[width=0.11\textwidth]{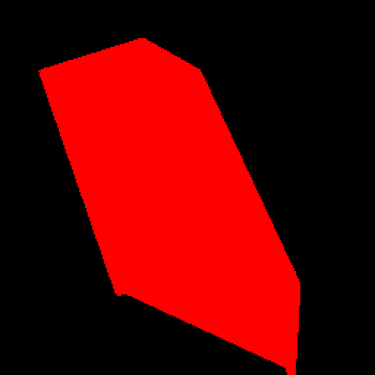} &
        \includegraphics[width=0.11\textwidth]{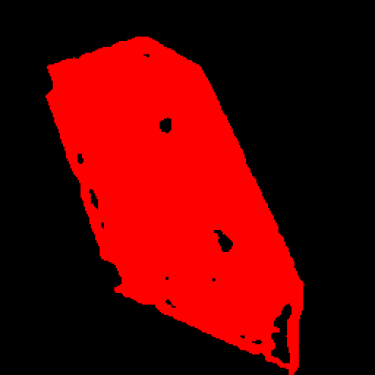} &
        \includegraphics[width=0.11\textwidth]{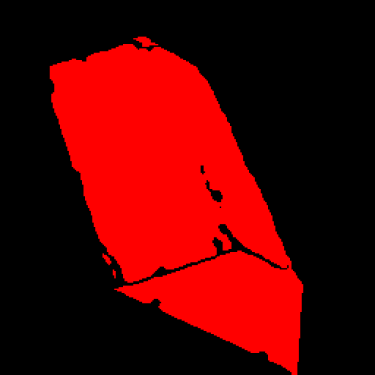} &
        \includegraphics[width=0.11\textwidth]{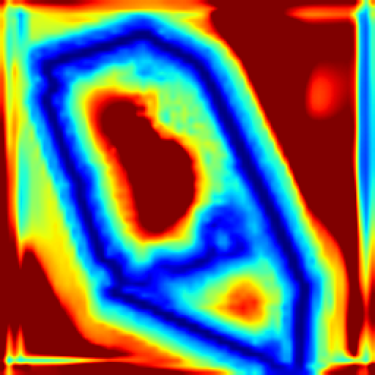} &
        \includegraphics[width=0.11\textwidth]{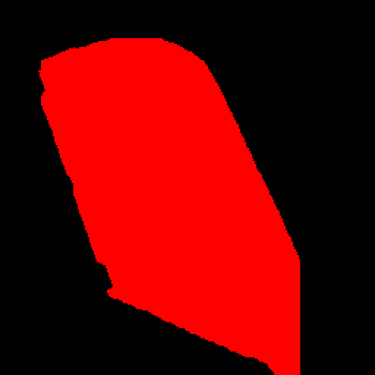} &
        \includegraphics[width=0.11\textwidth]{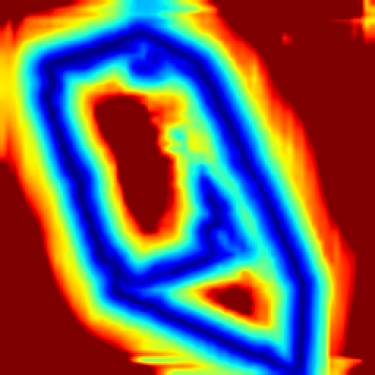} &
        \includegraphics[width=0.11\textwidth]{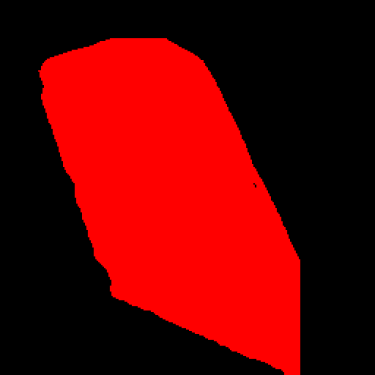} \\

        \includegraphics[width=0.11\textwidth]{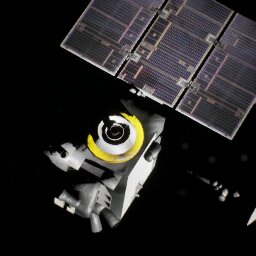} &
        \includegraphics[width=0.11\textwidth]{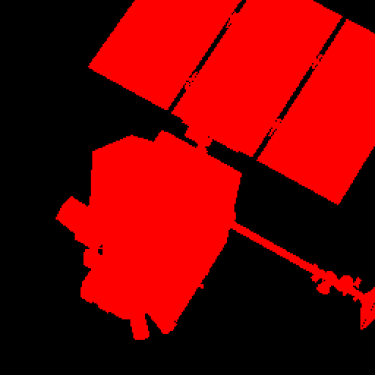} &
        \includegraphics[width=0.11\textwidth]{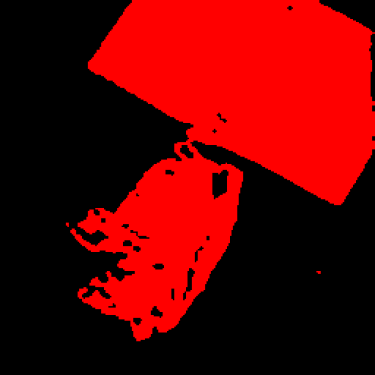} &
        \includegraphics[width=0.11\textwidth]{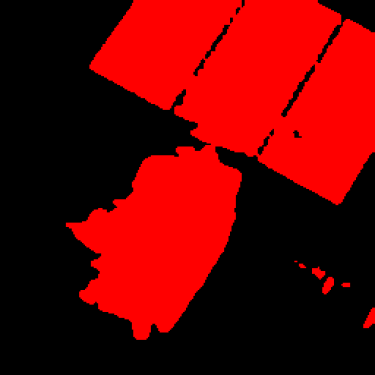} &
        \includegraphics[width=0.11\textwidth]{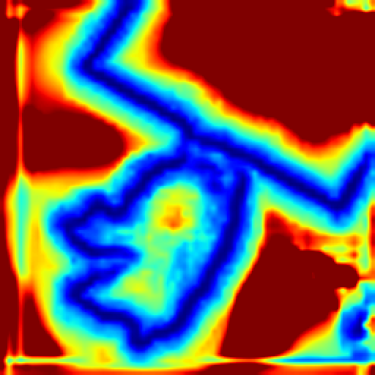} &
        \includegraphics[width=0.11\textwidth]{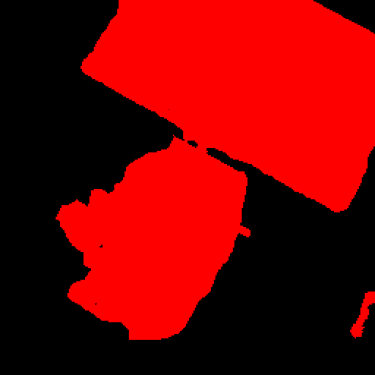} &
        \includegraphics[width=0.11\textwidth]{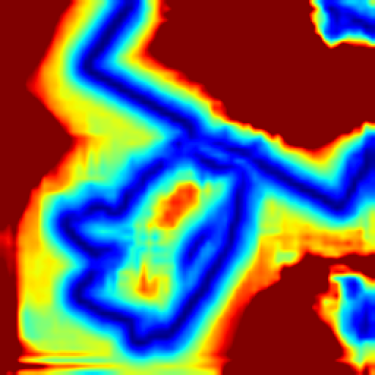} &
        \includegraphics[width=0.11\textwidth]{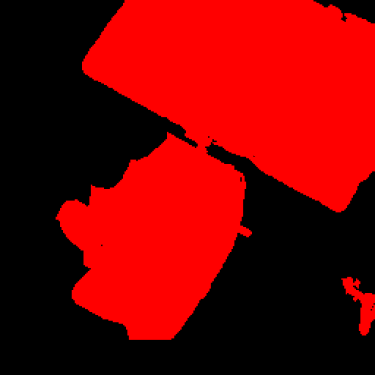} \\
    \end{tabular}

    \caption{Qualitative comparison across three spacecraft from SPE3R. Columns 1--2 show the RGB inputs and ground truth masks. Columns 3--4 represent the predictions of BL-v2 and SegFormer-b1. Columns 5--6 and 7--8 contain the predictions for distance field and segmentation mask of GABI-v3s and GABI-v2, respectively. Note that the proposed GABI models have substantially fewer parameters and FLOPs than the SegFormer counterparts(~\cref{tab:exp1}).}
    \label{fig:exp3}
\end{figure*}

\subsection{Generalization in SPARK}
For this experiment, we exclude the  \textit{Soho}  and \textit{Proba3ocs} spacecraft models from the training. These models were specifically selected due to their strong geometric distinctions from other spacecraft classes. The panels in \textit{Proba3ocs} resemble a combination of the structure of different models, whereas the \textit{Soho} spacecraft is labeled with panels only on one side of the panel frames. Furthermore, the Soho panels possess a distinct glow that is not observed to that extent in other spacecraft models.

~\cref{tab:exp2} reports the segmentation performance on the two hidden spacecraft models. All methods achieve relatively strong performance on the body class, while performance degrades at higher IoU thresholds for \textit{Soho}, mainly due to panel misclassification. In this challenging scenario, GABI achieves a substantial improvement in mAP $30$ - $45\%$ compared to the competing models.

These results suggest that although segmentation performance remains stable at lower IoU thresholds due to the homogeneous texture of the dataset, the distance-field supervision enables more precise boundary localization. Similar to the observations in ~\cref{sec:exp1}, SegFormer architectures exhibit limited performance in this setting.

The qualitative comparison in ~\cref{fig:exp2} offers the results of a case in which intense lens glare affects a portion of the panel. We can see that all heavier models have similar performance in panel identification, with GABI better predicting the shape and, hence, the mask of the spacecraft body. Notably, in the top row, where the lighter networks are depicted, we observe that GABI leverages the geometry learned from the distance-field prediction to strongly outperform the competitors and match the performance of the heavier networks. It is evident that the distance field enables smoother, spatially connected prediction masks.

\subsection{Cross-domain Generalization in SPE3R}
To test generalization across domains, we employ SPE3R, a dataset significantly different in texture and geometry. The evaluation results on three previously unseen spacecraft models: Apollo–Soyuz, Grace, and lro are summarized in~\cref{tab:exp3}.
Across all spacecraft models, GABI consistently outperforms its convolutional counterparts and achieves performance competitive with the much heavier transformer-based architectures.
The most notable improvements occur at higher IoU thresholds. For example, on the Apollo spacecraft, GABI-v2 improves $\text{AP}_{75}$ from $15$-$20\%$ to $65$-$70\%$, while the transformer architectures achieve between $54$-$67\%$. Similar improvements are observed in Grace, where the Average Precision for high IoU is almost doubled and it is still $20\%$ higher than the heavier and best-performing SegFormer. These gains also translate into strong overall segmentation quality, as indicated by the consistent improvement in the F1-score.

~\cref{fig:exp3} shows that GABI achieves performance comparable to, and in some cases better than, the heavier SegFormer architecture while using only $10\%$--$30\%$ of its complexity.
%
Geometric supervision allows GABI to produce less fragmented segmentation masks while learning complete boundaries, even in dim regions. The findings of the previous experiment about smooth and connected predictions translate across domains as well. Overall, the learned distance field appears sufficiently accurate for downstream applications.

\section{Conclusion}
In this work, we propose GABI, an efficient segmentation network augmented with explicit geometric supervision. We show that the distance field prediction head can successfully localize the boundaries of the segmentation masks and the total spacecraft by integrating structural information. The multi-task segmentation setup with boundary gating helps the network learn structural regularities across different spacecraft models. This geometric regularization improves generalization across datasets, often leading to performance increase comparable with much larger networks. The results also indicate that GABI-v2 and GABI-v3 consistently outperform their convolutional counterparts, with an improvement of up to $50$ -- $100\%$ AP in some cases.

The strong performance of the proposed approach opens several promising directions for future research. We plan to investigate the robustness of the predictions under varying illumination conditions and in the presence of severe visual artifacts. Finally, it is essential to assess the network's performance in a complete pipeline, in which GABI's outputs are used for downstream tasks such as pose estimation or tracking.

\section{Acknowledgments}
Iason Georgios Velentzas was supported by the AFOSR SURI project (FA9550-23-1-0723). Iason would like to thank George Rapakoulias for their insightful discussions and feedback that improved the quality of this work.

{
    \small
    \bibliographystyle{ieeenat_fullname}
    \bibliography{main}
}


\end{document}